\pdfoutput=1
\documentclass{article} 
\usepackage{iclr2017_conference_arxiv,times}
\usepackage{url}
\usepackage{color}
\usepackage[bookmarks=false]{hyperref}
\usepackage{kky}
\usepackage{graphicx} 
\usepackage{amsmath, amssymb, amsthm}
\usepackage{natbib}
\usepackage{algorithm}
\usepackage{algorithmic}
\usepackage{subfigure}
\usepackage{times}
\usepackage{bbm}
\usepackage{tikz}
\usepackage{pgfplots}
\usepackage{todonotes}
\usepackage{tikz-3dplot}

\newcommand{\ie}[0]{\emph{i.e.},~}
\newcommand{\eg}[0]{\emph{e.g.},~}

\title{Annealing Gaussian into ReLU: a New Sampling Strategy for Leaky-ReLU RBM}

\author{Chun-Liang Li\hspace{10pt} Siamak Ravanbakhsh\hspace{10pt} Barnab{\'{a}}s P{\'{o}}czos \\ 
Department of Machine Learning\\
Carnegie Mellon University\\
Pittsburgh, PA 15213, USA \\
\texttt{\{chunlial,mravanba,bapoczos\}@cs.cmu.edu} \\
}

\begin{document}

\maketitle

\begin{abstract}
Restricted Boltzmann Machine (RBM) is a bipartite graphical model that is used as the building block in energy-based
deep generative models. Due to numerical stability and quantifiability of the likelihood,  RBM is commonly used with
Bernoulli units. Here, we consider an alternative member of exponential family RBM with leaky rectified linear units --
called leaky RBM.  We first study the joint and marginal distributions of leaky RBM under different leakiness, which
provides us important insights by connecting the leaky RBM model and truncated Gaussian distributions.  The connection
leads us to a simple yet efficient method for sampling from this model, where the basic idea is to anneal the leakiness
rather than the energy; -- \ie start from a fully Gaussian/Linear unit and gradually decrease the leakiness over
iterations.  This serves as an alternative to the annealing of the temperature parameter and enables numerical
estimation of the likelihood that are more efficient and more accurate than the commonly used annealed importance
sampling (AIS).  We further demonstrate that the proposed sampling algorithm enjoys faster mixing property than
contrastive divergence algorithm, which benefits the training without any additional computational cost.
\end{abstract}

\section{Introduction}

In this paper, we are interested in deep generative models.  There is a family of directed deep generative models which
can be trained by back-propagation~\citep[\eg][]{KingmaW13, Goodfellow14}.  The other family is the deep energy-based models,
including deep belief network~\citep{Hinton2006} and deep Boltzmann machine~\citep{SalHinton07}. The building block of
deep energy-based models is a bipartite graphical model called restricted Boltzmann machine (RBM). The RBM model
consists of two layers, visible and hidden layers, which can model higher-order correlation of the visible units
(visible layer) using the hidden units (hidden layer). It also makes the inference easier that there are no interactions between the variables in each layer.  

The conventional RBM uses Bernoulli units for both the hidden and visible units~\citep{Smolensky1986}. One extension is using
Gaussian visible units to model general natural images~\citep{Freund1994}.  For hidden units, we can also generalize
Bernoulli units to the exponential family~\citep{WellingRH04, RavanbakhshPSSG16}.  

\cite{NairH10} propose one special case by using Rectified Linear Unit (ReLU) for the hidden layer with the heuristic
sampling procedure, which has promising performance in terms of reconstruction error and classification accuracy.
Unfortunately, due to its lack of strict monotonicity, ReLU RBM does not fit within the framework of exponential family
RBMs~\citep{RavanbakhshPSSG16}. Instead we study leaky-ReLU RBM (leaky RBM) in this work and address two important issues i) a
better training (sampling) algorithm for ReLU RBM and; ii) a better quantification of leaky RBM --\ie its
performance in terms of likelihood.

We study some of the fundamental properties of leaky RBM, including its joint and marginal
distributions (Section~\ref{sec:rbm}).  By analyzing these distributions, we show that the leaky RBM is a \emph{union of
truncated Gaussian distributions}.
In this paper we
will show that  training leaky RBM involves underlying positive definite constraints. Because of this, the
training can diverge if these constrains are not satisfied.
This is an issue that was previously ignored in ReLU RBM, as it was mainly used for pre-training rather than generative modeling.
Our contribution in this paper is three-fold:
I) We systematically identify and address model constraints in leaky RBM (Section~\ref{sec:relu});
II) For the training of leaky RBM, we propose a meta algorithm for sampling, which anneals leakiness during the Gibbs sampling procedure
(Section~\ref{sec:relu}) and empirically show that it can boost contrastive divergence with faster mixing (Section~\ref{sec:mix}); 
III) We demonstrate the power of the proposed sampling algorithm on estimating the partition
function. In particular, comparison on several benchmark datasets shows that the proposed method outperforms the
conventional AIS~\citep{salakhutdinov2008} (Section~\ref{sec:par}). 
Moreover, we provide an incentive for using leaky RBM by showing that the leaky
ReLU hidden units perform better than the Bernoulli units in terms of the model log-likelihood (Section~\ref{sec:par}).

\section{Restricted Boltzmann Machine and ReLU}
\label{sec:rbm}

The Boltzmann distribution 
is defined as
\[
	p(x) = \frac{e^{-E(x)}}{Z},
\]
where $Z=\sum_x e^{-E(x)}$ is the partition function.
Restricted Boltzmann Machine (RBM) is a Boltzmann distribution with a bipartite structure
It is also the building block for many deep models~\citep[\eg][]{Hinton2006, SalHinton07, Lee2009}, 
which are widely used in numerous applications~\citep{Bengio2009}. 
The conventional \emph{Bernoulli} RBM, models the joint probability $p(v,h)$ for the visible units $v\in[0,1]^I$ and the hidden units $h\in [0,1]^J$ as $p(v,h)\propto \exp(-E(v,h))$, where
\[
	E(v,h) = a^\top v - v^\top Wh + b^\top h.
\]
The parameters are $a\in \RR^I$, $b\in \RR^J$ and $W\in\RR^{I\times J}$.
We can derive the  conditional probabilities as
\begin{equation}
	p(v_i=1|h) = \sigma\left( \sum_{j=1}^J W_{ij}h_j + a_i \right)\hspace{10pt}\mbox{and}\hspace{10pt}p(h_j=1|v) = \sigma\left( \sum_{i=1}^I W_{ij}v_i + b_j \right),
	\label{eq:cond_b}
\end{equation}
where $\sigma(x) = (1 + e^{-x})^{-1}$ is the sigmoid function.  

One extension of Bernoulli RBM is replacing the binary visible units by linear units $v\in\RR^I$ with independent Gaussian noise. The energy function in this case is given by
\[
	E(v,h) = \sum_{i=1}^I \frac{(v_i-a_i)^2}{2\sigma_i^2} - \sum_{i=1}^I\sum_{j=1}^J \frac{v_i}{\sigma_i}W_{ij}h_j + b^\top h.
\]
To simplify the notation, we eliminate $a_i$ and $\sigma_i$ in this paper, and then the energy function is simplified to be 
$E(v,h) = \frac{\|v\|^2}{2} -  v^\top Wh + b^\top h$. Note that the elimination does not influence the discussion and
one can easily extend all the results in this paper to the model that includes $a_i$ and~$\sigma_i$. 

The conditional distributions are as follows:
\begin{equation}
	p(v_i|h) = \Ncal\left( \sum_{j=1}^J W_{ij}h_j, 1\right)\hspace{10pt}\mbox{and}\hspace{10pt}p(h_j=1|v) = \sigma\left( \sum_{i=1}^I W_{ij}v_i + b_j \right),
	\label{eq:cond_g}
\end{equation}
where $\Ncal(\mu, V)$ is a Gaussian distribution with mean $\mu$ and variance $V$.

\subsection{ReLU RBM with Continuous Visible Units}

From~\eqref{eq:cond_b} and~\eqref{eq:cond_g}, we can see that the mean of the $p(h_j|v)$ is
actually the evaluation of a sigmoid function at the response 
$\sum_{i=1}^I W_{ij}v_i + b_j $, which is the non-linearity of the hidden units. From this perspective, we can extend the sigmoid function to other
functions and thus allow RBM to have more expressive power~\citep{RavanbakhshPSSG16}.
\cite{NairH10} propose to use \emph{rectified linear unit} (ReLU) to replace
conventional sigmoid hidden units.  The activation function is defined as a
one-sided function $\max(0, x)$. 

However, as it has been shown in~\cite{RavanbakhshPSSG16}, only the strictly monotonic activation functions can derive feasible joint and conditional
distributions\footnote{\cite{NairH10} use the heuristic noisy ReLU for sampling.}. Therefore, we consider the leaky ReLU~\citep{maas2013rectifier} in this paper. 
The activation function of leaky ReLU is defined as $\max(cx, x)$, where $c\in (0,1)$ is the leakiness parameter. 

To simplify the notation, we define $\eta_j = \sum_{i=1}^I W_{ij}v_i + b_j$.
By~\cite{RavanbakhshPSSG16}, 
the conditional probability of the activation $f$ is defined as $p(h_j|v) = \exp\left(-D_f(\eta_j\|h_j) + g(h_j) \right)$, where $D_f(\eta_j\|h_j)$ is a 
Bregman Divergence and $g(h_j)$ is the base measure.
The Bergman divergence of $f$ is given by $D_f(\eta_j\|h_j) = -\eta_jh_j + F(\eta_j) + F^*(h_j)$, where $F$ with $\frac{d}{d\eta_j}F(\eta_j) = f(\eta_j)$ 
is the anti-derivative of $f$ and $F^*$ is the anti-derivative of $f^{-1}$.
We then get the conditional distributions of leaky RBM as 
\begin{equation}
	p(h_j|v) = 
	\begin{cases}
		\Ncal(\eta_j, 1), & \text{if } \eta_j > 0 \\
		\Ncal(c\eta_j, c), & \text{if } \eta_j \leq 0.
	\end{cases}
	\label{eq:hv}
\end{equation}
Note that the conditional distribution of the visible unit is  
\begin{equation}
	p(v_i|h) = \Ncal\left( \sum_{j=1}^J W_{ij}h_j, 1\right),
	\label{eq:vh}
\end{equation}
which can also be written as $p(v_i|h) = \exp\left(-D_{\tilde{f}}(\nu_i\|v_i) + g(v_i)\right)$, where $\nu_i = \sum_{j=1}W_{ij}h_j$ and $\tilde{f}(x) =x$. 
By having these two conditional distributions, we can train and do inference on a leaky RBM model by using contrastive
divergence~\citep{Hinton2002a} or other 
algorithms~\citep{Tieleman2008, tielemanFpcd2009}.

\section{Training and Sampling from leaky RBM}
\label{sec:relu}
First, we explore the joint and marginal distribution of the leaky RBM.
Given the conditional distributions $p(v|h)$ and $p(h|v)$, the joint distribution $p(v,h)$ from the general treatment for MRF model given by \cite{YangRAL12} is
\begin{equation}
	p(v,h) \propto \exp\left( v^\top Wh - \sum_{i=1}^I(\tilde{F}^*(v_i)+g(v_i)) - \sum_{j=1}^J(F^*(h_j)+g(h_j))\right).
	\label{eq:joint_gen}
\end{equation}
By~\eqref{eq:joint_gen}, we can derive the joint distribution of the leaky-ReLU RBM as
\[
	p(v, h) \propto \exp\left( v^\top Wh - \frac{\|v\|^2}{2} - \sum_{\eta_j>0}\left(\frac{h_j^2}{2}+\log\sqrt{2\pi}\right) 
	- \sum_{\eta_j\leq 0}\left(\frac{h_j^2}{2c}+\log\sqrt{2c\pi}\right) + b^\top h
	\right),
\]
and the marginal distribution as
\begin{equation}
	\begin{array}{ccl}
		p(v) & \propto & \displaystyle\exp\left( -\frac{\|v\|^2}{2} \right)\prod_{\eta_j>0}\exp\left( \frac{\eta_j^2}{2} \right) \prod_{\eta_j\leq 0}\left( \frac{c\eta_j^2}{2}\right) \\
			& \propto & \displaystyle\exp\left( -\frac{1}{2}v^\top\left( I-\sum_{\eta_j>0}W_jW_j^\top - c\sum_{\eta_j\leq 0}W_jW_j^\top \right)v + \sum_{\eta_j>0}b_jW_j^\top v + c\sum_{\eta_j\leq 0}b_jW_j^\top v \right). \\
	\end{array}
	\label{eq:joint}
\end{equation}
where $W_j$ is the $j$-th column of $W$.

\subsection{Leaky RBM as Union of Truncated Gaussian Distributions}
\label{sec:truncate}
From \eqref{eq:joint}, the marginal probability is determined by the affine constraints $\eta_j>0$ or $\eta_j\leq 0$ for all $j$.  
By combinatorics, these constraints divide $\RR^I$ into at most $M = \sum_{i=1}^I \binom Ji$ convex regions $R_1,
\cdots R_{M}$. An example with $I=2$ and $J=3$ is shown in Figure~\ref{fig:toy}.  If $I>J$, then we have at most $2^J$ regions.

\begin{figure}
	\begin{minipage}[b]{0.32\linewidth}
		\centering
		\usetikzlibrary{arrows,intersections}

\begin{tikzpicture}[scale=0.32]
    \draw (0,0) -- (6,6)  ; 
    \draw (2,6) -- (8,0)  ; 
	\draw (-1, 1.5) -- (9, 1.5)  ; 
	 \node [draw=none, color=red] at (6.3,6.3) {\normalsize $W_1$};
	 \node [draw=none, color=red] at (8.5,-0.4) {\normalsize $W_2$};
	 \node [draw=none, color=red] at (9.6,1.6) {\normalsize $W_3$};
	 \node [draw=none] at (4,2.5) {\small $R_1$};
	 \node [draw=none] at (4,0) {\small $R_2$};
	 \node [draw=none] at (8.5,0.75) {\small $R_3$};
	 \node [draw=none] at (8,4) {\small $R_4$};
	 \node [draw=none] at (4,5.6) {\small $R_5$};
	 \node [draw=none] at (0,4) {\small $R_6$};
	 \node [draw=none] at (-0.5,0.75) {\small $R_7$};
\end{tikzpicture}
		\caption{A two dimensional example with $3$ hidden units.}
		\label{fig:toy}
	\end{minipage}
	\hspace{5pt}
	\begin{minipage}[b]{0.32\linewidth}
		\centering
		\pgfmathdeclarefunction{gauss}{2}{%
  \pgfmathparse{1/(#2*sqrt(2*pi))*exp(-((x-#1)^2)/(2*#2^2))}%
}

\begin{tikzpicture}[scale=0.4]
\begin{axis}[every axis plot post/.append style={
	mark=none,domain=-1:1,samples=50,smooth}, 
	axis x line*=bottom, 
	axis y line*=left, 
	ymin=0,
	xtick={4,6.5}
	] 
	\addplot[very thick, blue] {gauss(0,0.7)};
	\addplot[very thick, red] {gauss(0,1)};
	\draw[dashed] (0, 0) -- (0,250) node[below] {};
	\draw[dashed] (200, 0) -- (200,250) node[below] {};
	\end{axis}
\end{tikzpicture}
		\caption{An one dimensional example of truncated Gaussian distributions with different variances.}
		\label{fig:toy}
	\end{minipage}
	\hspace{5pt}
	\begin{minipage}[b]{0.32\linewidth}
		\centering
		\tdplotsetmaincoords{70}{90}
\begin{tikzpicture}[scale=1,tdplot_main_coords,>=latex, x={(1,-0.5,0)}]
	\draw[thick,->] (-2,0,0)--(2,0,0) node[anchor=north east]{$x$};
	\draw[->, thick] (0,-2,0)--(0,2,0) node[anchor=north east]{$y$};
	\draw[thick,->] (0,0,-1.5)--(0,0,1.5) node[anchor=north east]{$z$};
	\fill[blue, opacity=0.3] (-1.5,0,-0.5) -- (-1.5,0,0.5) -- (1.5,0,0.5) -- (1.5,0,-0.5) --cycle;
	\fill[blue, opacity=0.3] (0.5,-1.5,0) -- (-0.5,-1.5,0) -- (-0.5,1.5,0) -- (0.5,1.5,0) --cycle;
	\fill[blue, opacity=0.3] (0,0.5,-1.2) -- (0,-0.5,-1.2) -- (0,-0.5,1.2) --(0,0.5,1.2) -- cycle;
	\node at (0,1.5,1) {\large$\mathbb{R}^3$};
	\node at (1.6,0,-0.5) {\normalsize$W_1$};
	\node at (0.5,1.6,0) {\normalsize$W_2$};
	\node at (0,0.5,1.3) {\normalsize$W_3$};
\end{tikzpicture}
		\caption{A three dimensional example with $3$ hidden units, where $W_j$ are orthogonal to each other.}
		\label{fig:toy3}
	\end{minipage}
\end{figure}

We discuss the two types of these regions. For bounded regions, such as $R_1$ in Figure~\ref{fig:toy}, the integration
of~\eqref{eq:joint} is also bounded, which results in a valid distribution.  Before we discuss the unbounded cases, we define
$\Omega = I-\sum_{j=1}^J\alpha_jW_jW_j^\top$, where $\alpha_j = \mathbbm{1}_{\eta_j>0} + c\mathbbm{1}_{\eta_j\leq 0}$. 
For the unbounded region, 
if $\Omega \in \RR^{I\times I}$ is a positive definite (PD) matrix, then the probability density is proportional to a multivariate Gaussian distribution
with mean $\mu=\Omega^{-1}\left( \sum_{j=1}^J\alpha_jb_jW_j \right)$ and precision matrix $\Omega$ (covariance matrix
$\Omega^{-1}$) but over an affine-constrained region.
Therefore, the distribution of each unbounded region can be
treated as a truncated Gaussian distribution. 

On the other hand, if $\Omega$ is not PD, and the region
$R_i$ contains the eigenvectors with negative eigenvalues of $\Omega$, the integration of \eqref{eq:joint} over $R_i$ is
divergent (infinite), which can not result in a valid probability distribution. In practice, with this type of parameter, when we do Gibbs sampling on the
conditional distributions, the sampling will diverge.  However, it is unfeasible to check exponentially many regions for each gradient
update.

\begin{theorem}
\label{thm:upper}
If $I-WW^\top$ is positive definite, then $I-\sum_j \alpha_jW_jW_j^\top $ is also positive definite, for all $\alpha_j \in [0,1]$.
\end{theorem}
The proof is shown in Appendix~\ref{thm:upper}.
From Theorem~\ref{thm:upper} we can see that if the constraint $I-WW^\top$ is PD, then one can guarantee that the distribution of every region is a valid truncated Gaussian distribution. 
Therefore, we introduce the following projection step for each $W$ after the gradient update.
\begin{equation}
	\begin{array}{cl}
		\displaystyle \argmin_{\tilde{W}} & \|W-\tilde{W}\|_F^2 \\
		\mbox{s.t.} & I-\tilde{W}\tilde{W}^\top \succeq 0
	\end{array}
	\label{eq:proj}
\end{equation}

\begin{theorem}
\label{thm:proj}
The above projection step~\eqref{eq:proj} can be done by shrinking the singular values to be less than 1. 
\end{theorem}
The proof is shown in Appendix~\ref{sec:pf}.
The training algorithm of the leaky RBM is shown in Algorithm~\ref{algo:train}.
By using the projection step \eqref{eq:proj}, we could treat the leaky RBM as the \emph{union of truncated Gaussian
distributions}, which uses weight vectors to divide the space of visible units into several regions and use a truncated
Gaussian distribution to model each region. Note that the leaky RBM model is different from~\cite{SuLCC16}, which uses a truncated Gaussian
distribution to model the conditional distribution $p(h|v)$ instead of the marginal distribution.
The empirical study about the divergent values and the necessity of the projection step is shown in Appendix~\ref{sec:div}.
\begin{algorithm}
  \begin{algorithmic} 
    \FOR{$t=1,\ldots, T$}
      \STATE Estimate gradient $g_\theta$ by CD or other algorithms with ~\eqref{eq:hv} and~\eqref{eq:vh}, where $\theta
	  = \{W, a, b\}$.
	  \STATE $\theta^{(t)} \leftarrow \theta^{(t-1)} + \eta g_\theta$.
      \STATE Project $W^{(t)}$  by~\eqref{eq:proj}.
    \ENDFOR
    \caption{Training Leaky RBM}
    \label{algo:train}
  \end{algorithmic}
\end{algorithm}

\subsection{Sampling from Leaky-ReLU RBM}
Gibbs sampling is the core procedure for RBM, including
training, 
inference, 
and estimating the partition function~\citep{FischerI12,Tieleman2008,salakhutdinov2008}.  For every task,
we start from randomly initializing $v$ by an arbitrary distribution~$q$, and
iteratively sample from the conditional distributions. Gibbs sampling
guarantees the procedure result in the stationary distribution in the long run for any initialized distribution $q$.
However, if $q$ is close to the target distribution $p$, it can significantly shorten the number of iterations to achieve the stationary distribution.

If we set the leakiness $c$ to be $1$, then \eqref{eq:joint} becomes a simple multivariate Gaussian distribution $\Ncal\left( (I-WW^\top)^{-1}Wb, (I-WW^\top)^{-1}\right)$, which 
can be easily sampled without Gibbs sampling. Also, the projection step~\eqref{eq:proj} guarantees it is a valid Gaussian distribution.
Then we decrease the leakiness with a small $\epsilon$, and use samples from the multivariate Gaussian distribution when $c=1$ as the initialization to do Gibbs sampling. 
Note that the distribution of each region is a truncated Gaussian distribution. 
When we only decrease the leakiness with a small amount, the resulted distribution is a ``similar'' truncated Gaussian 
distribution with more concentrated density. 
From this observation, we could expect the original multivariate Gaussian distribution serves as a good initialization.
The one-dimensional example is shown in Figure~\ref{fig:toy}.
We then repeat this procedure until we reach the target leakiness. The algorithm can be seen as \emph{annealing the leakiness} during the Gibbs sampling procedure. 
The meta algorithm is shown in Algorithm~\ref{algo:sample}.
Next, we show the proposed sampling algorithm can help both the partition function estimation and the training of leaky RBM.

\begin{algorithm}
  \begin{algorithmic} 
  	\STATE Sample $v$ from $\Ncal\left( (I-WW^\top)^{-1}Wb, (I-WW^\top)^{-1}\right)$
	\STATE $c'=1$
    \FOR{$t=1,\ldots, T$}
		\IF{$c'>c$ }  
			\STATE $c' = c'-\epsilon$
		\ENDIF
		\STATE Do Gibbs sampling by using ~\eqref{eq:hv} and~\eqref{eq:vh} with leakiness $c'$
    \ENDFOR
    \caption{Meta Algorithm for Sampling from Leaky RBM}
    \label{algo:sample}
  \end{algorithmic}
\end{algorithm}

\section{Partition Function Estimation}
\label{sec:par}
It is known that estimating the partition function of RBM is intractable~\citep{salakhutdinov2008}. 
Existing approaches, including~\cite{salakhutdinov2008, GrosseMS13, liu2015, Carlson16} focus on using sampling 
to approximate the partition function of the conventional Bernoulli RBM instead of the RBM with Gaussian visible units
and non-Bernoulli hidden units.
In this paper, we focus on extending the classic annealed importance sampling (AIS) algorithm~\citep{salakhutdinov2008}
to leaky RBM.

Assuming that we want to estimate the partition function $Z$ of $p(v)$ with $p(v)=p^*(v)/Z$ and  $p^*(v) \propto \sum_h \exp( -E(v,h) )$, 
\cite{salakhutdinov2008} start from a initial distribution $p_0(v) \propto \sum_h \exp( -E_0(v,h) )$, where computing the partition $Z_0$ of $p_0(v)$ is tractable and we can draw samples from $p_0(v)$.
They then use the ``geometric path'' to anneal the intermediate distribution as $p_k(v) \propto p_k^*(v) = \sum_h \exp\left( -\beta_kE_0(v,h) - (1-\beta_k)E(v,h) \right)$, 
where they grid $\beta_k$ from $1$ to $0$. 
If we let $\beta_0=1$, we can draw samples $v_k$ from $p_k(v)$ by using samples $v_{k-1}$ from $p_{k-1}(v)$ for $k\geq 1$
via Gibbs sampling.
The partition function is then estimated via $Z$ = $\frac{Z_0}{M}\sum_{i=1}^M \omega^{(i)}$, where 
\[
	\omega^{(i)} = \frac{ p_1^*(v_0^{(i)}) }{ p_0^*(v_0^{(i)}) }\frac{ p_2^*(v_1^{(i)}) }{ p_1^*(v_1^{(i)}) }\cdots\frac{ p_{K-1}^*(v_{K-2}^{(i)}) }{ p_{K-2}^*(v_{K-2}^{(i)}) }\frac{ p_{K}^*(v_{K-1}^{(i)}) }{ p_{K-1}^*(v_{K-1}^{(i)}) },
\]
and $\beta_K=0$.

\cite{salakhutdinov2008} use the initial distribution with independent visible units and without hidden units. 
Therefore, we extend~\cite{salakhutdinov2008} to the leaky-ReLU case with $E_0(v,h) = \frac{\|v\|^2}{2}$, which results
in a multivariate Gaussian distribution $p_0(v)$.
Compared with the meta algorithm shown in Algorithm~\ref{algo:sample} which \emph{anneals between leakiness}, 
the extension of \cite{salakhutdinov2008} \emph{anneals between energy functions}.

\subsection{Study on Toy Examples}
As we discussed in Section~\ref{sec:truncate}, leaky RBM with $J$ hidden units is a union of $2^J$ truncated Gaussian distributions.
Here we perform a study on the leaky RBM with a small number hidden units. Since in this example the number of hidden units is small, we can integrate out all possible configurations of $h$.
However, integrating a truncated Gaussian distribution with general affine constraints does not have analytical
solutions, and
several approximations have been developed~\citep[\eg][]{Pakman14}. To compare our results with the exact partition function, we consider a special case that has the following form:
\begin{equation}
	p(v) \propto \displaystyle\exp\left( -\frac{1}{2}v^\top\left( I-\sum_{\eta_j>0}W_jW_j^\top - c\sum_{\eta_j\leq 0}W_jW_j^\top \right)v  \right). 
	\label{eq:special}
\end{equation}
Compared to \eqref{eq:joint}, it is equivalent to the setting where $b=0$.
Geometrically, every $W_j$ passes through the origin. 
We further put the additional constraint $W_i\perp W_j, \forall i \neq j$. Therefore. we divide the whole space into $2^J$ equally-sized regions. 
A three dimensional example is shown in Figure~\ref{fig:toy3}.
Then the partition function of this special case has the analytical form 
\[
	Z = \frac{1}{2^J}\sum_{\alpha_j\in\{1,c\}, \forall j} (2\pi)^{-\frac{I}{2}}\left|\left(I-\sum_{j=1}^J\alpha_jW_jW_j^\top\right)^{-\frac{1}{2}}\right|.
\]

We randomly initialize $W$ and use SVD to make each column orthogonal to each other.
Also, we scale $\|W_j\|$ to satisfy $I-WW^\top \succeq 0$.
The leakiness parameter is set to be
$0.01$.  For ~\cite{salakhutdinov2008} (AIS-Energy), we use $10^5$ particles with $10^5$ intermediate distributions. For
the proposed method (AIS-Leaky), we use only $10^4$ particles with $10^3$ intermediate distributions. In this small problem we study the
cases when the model has $5, 10, 20$ and $30$ hidden units and $3072$ visible units.  The true log partition function $\log Z$ is shown in
Table~\ref{tb:partition} and the difference between $\log Z$ and the estimates given by the two algorithms are shown in
Table~\ref{tb:diff}.
\begin{table}
	\centering
	\begin{tabular}{c|c|c|c|c}
		\hline
		& $J=5$ & $J=10$ & $J=20$ & $J=30$ \\
		\hline
		Log partition function & $2825.48$ & $2827.98$ & $2832.98$ & $2837.99$ \\
		\hline
	\end{tabular}
	\caption{The true partition function for Leaky-ReLU RBM with different number of hidden units.}
	\label{tb:partition}
	\vspace{-5pt}
\end{table}

\begin{table}
	\centering
	\begin{tabular}{c|c|c|c|c}
		\hline
		& $J=5$ & $J=10$ & $J=20$ & $J=30$ \\
		\hline
		AIS-Energy & $1.76\pm 0.011$ & $3.56\pm 0.039$ & $7.95\pm 0.363$ & $9.60\pm 0.229$\\
		\hline
		AIS-Leaky & $\mathbf{0.02\pm 0.001}$ & $\mathbf{0.04 \pm 0.002}$& $\mathbf{0.08\pm 0.003}$ & $\mathbf{0.13\pm 0.004}$ \\
		\hline
	\end{tabular}
	\caption{The difference between the true partition function and the estimations of two algorithms with standard deviation.}
	\label{tb:diff}
	\vspace{-5pt}
\end{table}

From Table~\ref{tb:partition}, we observe that AIS-Leaky has significantly better and more stable estimations than
AIS-Energy especially when $J$ is large.  For example, when we increase $J$ from $5$ to $30$, the bias (difference) of
AIS-Leaky only increases from $0.02$ to $0.13$; however, the bias of AIS-Energy increases from $1.76$ to $9.6$. Moreover, we note
that AIS-Leaky uses less particles and less intermediate distributions, and therefore is more computationally efficient
than AIS-Energy. We further study the implicit connection between the proposed AIS-Leaky and AIS-Energy in
Appendix~\ref{sec:equiv}, which shows AIS-Leaky is a special case of AIS-Energy under certain conditions.

\subsection{Comparison between leaky-ReLU RBM and Bernoulli-Gaussian RBM}
It is known that the reconstruction error is not a proper approximation of the likelihood~\citep{Hinton12}.
By having an accurate estimation of the partition function, we can study the power of leaky RBM when our goal is to use under the likelihood function as our objective instead of the reconstruction error.

We compare the Bernoulli-Gaussian RBM\footnote{Our GPU implementation with gnumpy and cudamat can reproduce the results
of http://www.cs.toronto.edu/~tang/code/GaussianRBM.m}, which has Bernoulli hidden units and
Gaussian visible units.  We trained both models with CD-20\footnote{CD-n means that contrastive divergence was run for n
steps} and momentum.  For both model, we all used $500$ hidden units. We
initialized $W$ by sampling from $\text{Unif}(0, 0.01)$, $a=0$, $b=0$ and $\sigma=1$. The momentum parameter was $0.9$ and 
the batch size was set to $100$.
We tuned the learning rate between $10^{-1}$ and $10^{-6}$. We studied two benchmark data sets, including CIFAR10 and SVHN. The data
was normalized to have zero mean and standard deviation of~$1$ for each pixel.
The results of the log-likelihood values are reported in Table~\ref{tb:lld}.

\begin{table}
	\centering
	\begin{tabular}{c|c|c}
		\hline
		& CIFAR-10 & SVHN \\
		\hline
	Bernoulli-Gaussian RBM & $-2548.3$ & $-2284.2$ \\
		\hline
	Leaky-ReLU RBN & $-1031.1$ & $-182.4$ \\ 
		\hline
	\end{tabular}
	\caption{The log-likelihood performance of Bernoulli-Gaussian RBM and leaky RBM.}
	\label{tb:lld}
	\vspace{-5pt}
\end{table}

From Table~\ref{tb:lld}, leaky RBM outperforms Bernoulli-Gaussian RBM significantly. The unsatisfactory
performance of Bernoulli-Gaussian RBM may be in part due to the optimization procedure. If we tune the decay schedule of the learning-rate
for each dataset in an ad-hoc way, we observe the performance of Bernoulli-Gaussian RBM can be improved by 
$\sim 300$ nats 
for both datasets. Also, increasing CD-steps brings slight improvement. The other possibility
is the bad mixing during the CD iterations. The advanced algorithms~\cite{Tieleman2008, tielemanFpcd2009} may help.
Although~\cite{NairH10} demonstrate the power of ReLU in terms of reconstruction error and classification accuracy,
it does not imply its superior generative capability. 
\textbf{Our study confirms leaky RBM could have a much better generative performance compared to Bernoulli-Gaussian RBM}

\section{Better Mixing by Annealing Leakiness}
\label{sec:mix}

In this section, we show the idea of annealing between leakiness benefit the mixing in Gibbs sampling in other settings. A common procedure for comparison of sampling methods for RBM is through visualization. 
Here, we are interested in more quantitative metrics and the practical benefits of improved sampling. For this, we consider \emph{optimization performance} 
as the evaluation metric.  

The gradient of the log-likelihood function $\Lcal(\theta|v_{data})$ of general RBM models is
\begin{equation}
	\frac{\partial \Lcal(\theta|v_{data})}{\partial \theta} = \EE_{h|v_{data}}\left[ \frac{ \partial E(v,h) }{ \partial \theta }
	\right] -  \EE_{v,h}\left[ \frac{ \partial E(v,h) }{ \partial \theta } \right]. 
	\label{eq:lld}
\end{equation}
Since the second expectation in~\eqref{eq:lld} is usually intractable, people use different
algorithms~\citep{FischerI12} to
approximate it.

In this section, we compare two gradient approximation procedure. The first one is the conventional contrastive
divergence (CD)~\citep{Hinton2002a}.
The second method is using Algorithm~\ref{algo:sample} (Leaky) with the same number of mixing steps as CD. 
The experiment setup is the same as that of Section~\ref{sec:par}.
\begin{figure}
  \centering
  \subfigure[SVHN]{\includegraphics[width=0.45\textwidth]{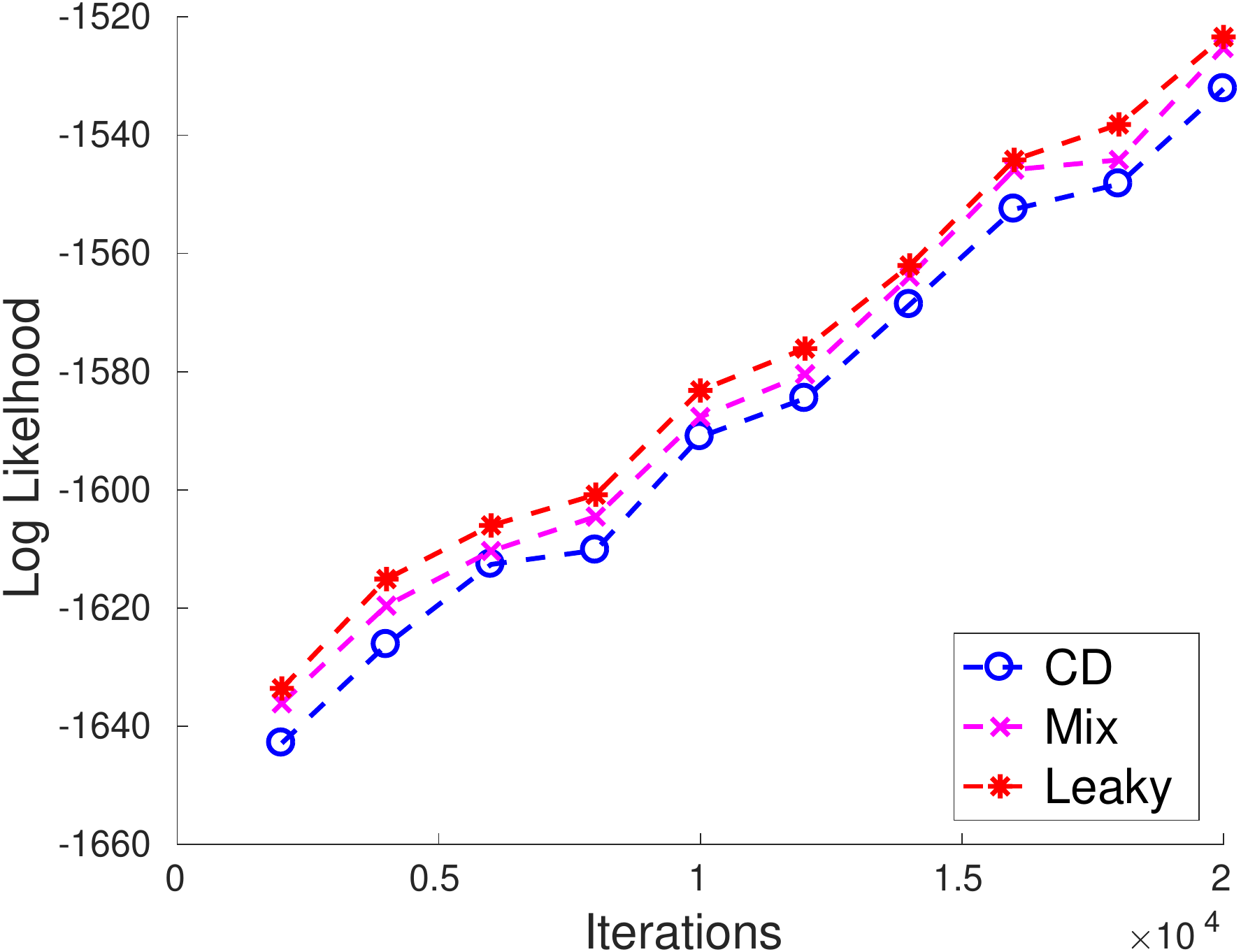}} 
  \subfigure[CIFAR10]{\includegraphics[width=0.45\textwidth]{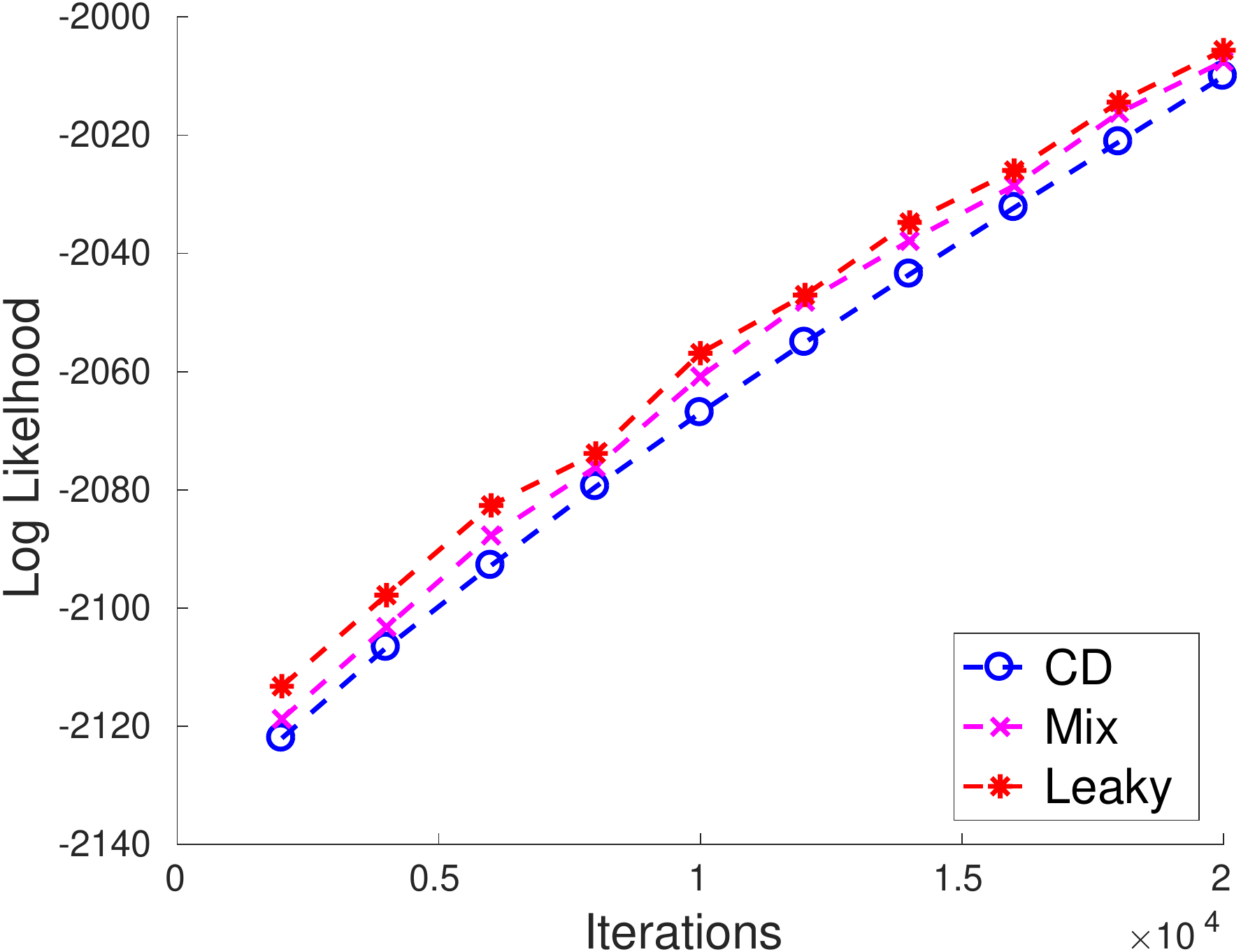}} 
  \caption{Training leaky RBM with different sampling algorithms.}
  \label{fig:opt}
  \vspace{-10pt}
\end{figure}
The results are shown in Figure~\ref{fig:opt}. 
The proposed sampling procedure is slightly better than typical CD steps. The reason is we only anneals the leakiness for
$20$ steps. To get accurate estimation requires thousands of steps as shown in Section~\ref{sec:par} when we
estimate the partition function. Therefore, the estimated gradient is still inaccurate. However, it still outperforms the 
conventional CD algorithm, which can demonstrate the better mixing power of the
proposed sampling algorithm as we expect. 

The drawback of using Algorithm~\ref{algo:sample} is sampling $v$ from $\Ncal\left( (I-WW^\top)^{-1}Wb, (I-WW^\top)^{-1}\right)$ 
requires computing mean, covariance and the Cholesky decomposition of the covariance matrix in every iteration, which are 
computationally expensive. We study a mixture algorithm by combining CD and the idea of annealing leakiness. 
The mixture algorithm is replacing the sampling $v$ from $\Ncal\left( (I-WW^\top)^{-1}Wb, (I-WW^\top)^{-1}\right)$ with
sampling from the empirical data distribution. The resulted mix algorithm is almost the same as CD algorithm while it
anneals the leakiness over the iterations as Algorithm~\ref{algo:sample}.
The results of the mix algorithm is also shown in Figure~\ref{fig:opt}.

The mix algorithm is slightly worse than the original leaky algorithm, but outperforms the conventional CD algorithm.
Starting from the data distribution is biased to $\Ncal\left( (I-WW^\top)^{-1}Wb, (I-WW^\top)^{-1}\right)$, which cause
the mix algorithm perform worse than Algorithm~\ref{algo:sample}.  However, by sampling from the data distribution, it
is as efficient as the CD algorithm (without additional computation cost).  Annealing the leakiness helps the mix
algorithm explore different modes of the distribution, which benefits the training. The idea could also be combined with
more advanced algorithms~\citep{Tieleman2008, tielemanFpcd2009}\footnote{We studied the PCD extension of the proposed
sampling algorithm. However, the performance is not as stable as CD.}.

\section{Conclusion}
In this paper, we study the properties of the distributions of leaky RBM. 
The study links the leaky RBM model and truncated Gaussian distributions. 
Also, our study shows and addresses an underlying positive definite constraint of training leaky RBM.
Based on our study, we further propose a meta sampling algorithm, which anneals between leakiness during the Gibbs sampling
procedure. 
We first demonstrate the proposed sampling algorithm is more effective and more efficient in estimating the partition function than the
conventional AIS algorithm. 
Second, we show the proposed sampling algorithm has better mixing property 
under the evaluation via optimization.

A few direction worth further studying. For example, one is how to speed up the naive projection step. Some potential
direction is using the barrier function as shown in~\cite{HSDP11} to avoid the projection step.

\bibliography{paper}
\bibliographystyle{iclr2017_conference}
\appendix
\section{Proof of Theorem~\ref{thm:upper}}
\begin{proof} 
Since $WW^\top - \sum_j \alpha_jW_jW_j = \sum_j (1-\alpha_j)W_jW_j^\top\succeq 0$, we have $WW^\top \succeq \sum_j \alpha_jW_jW_j$. 
Therefore, $I-\sum_j \alpha_jW_jW_j^\top \succeq I-WW^\top \succeq 0$.
\end{proof}

\section{Proof of Theorem~\ref{thm:proj}}
\label{sec:pf}
\begin{proof}
Let the SVD decomposition of $W$ and $\tilde{W}$ as $W=USV^\top$ and $\tilde{W}=\tilde{U}\tilde{S}\tilde{V}^\top$. Then we have
\begin{equation}
	\|W-\tilde{W}\|_F^2 = \| USV^\top-\tilde{U}\tilde{S}\tilde{V}^\top \|_F^2 \geq 
	\sum_{i=1}^I(S_{ii}-\tilde{S}_{ii})^2,
	\label{eq:svd}
\end{equation}
and the constraint $I-\tilde{W}\tilde{W}^\top \succeq 0$ can be rewritten as $0 \leq \tilde{S}_{ii} \leq 1, \forall i$. 
The transformed problem has a Lasso-like formulation and we can solve it by $\tilde{S}_{ii} = \min(S_{ii},
1)$~\citep{Parikh14}. Also, the lower
bound $\sum_{i=1}^I(S_{ii}-\tilde{S}_{ii})^2$ in \eqref{eq:svd} becomes a tight bound when we set $\tilde{U} = U$ and
$\tilde{V} = V$, which completes the proof.
\end{proof}

\section{Necessity of the Projection Step} 
\label{sec:div}
\begin{figure}[h]
  \centering
  \subfigure[SVHN]{\includegraphics[width=0.45\textwidth]{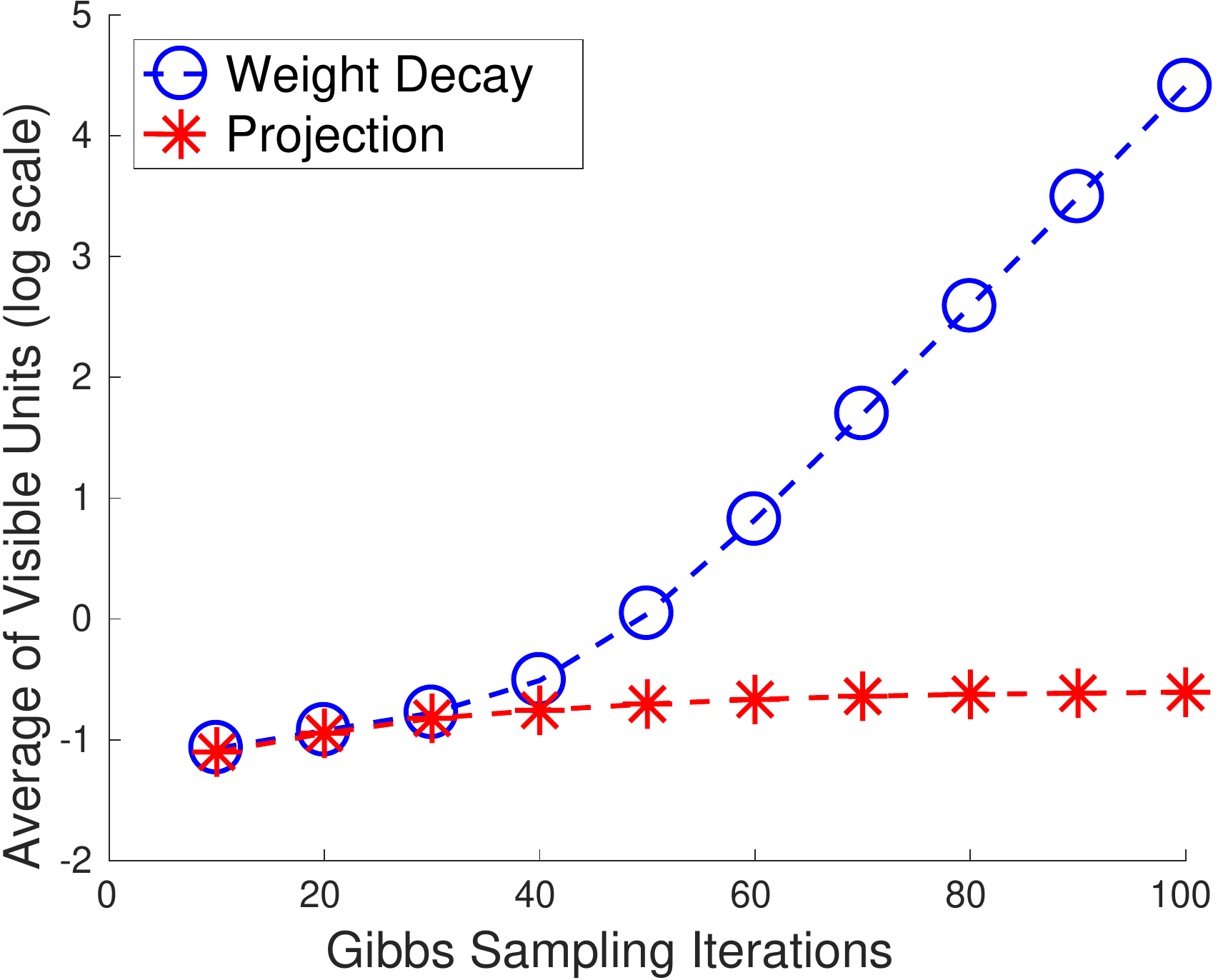}} 
  \subfigure[CIFAR10]{\includegraphics[width=0.45\textwidth]{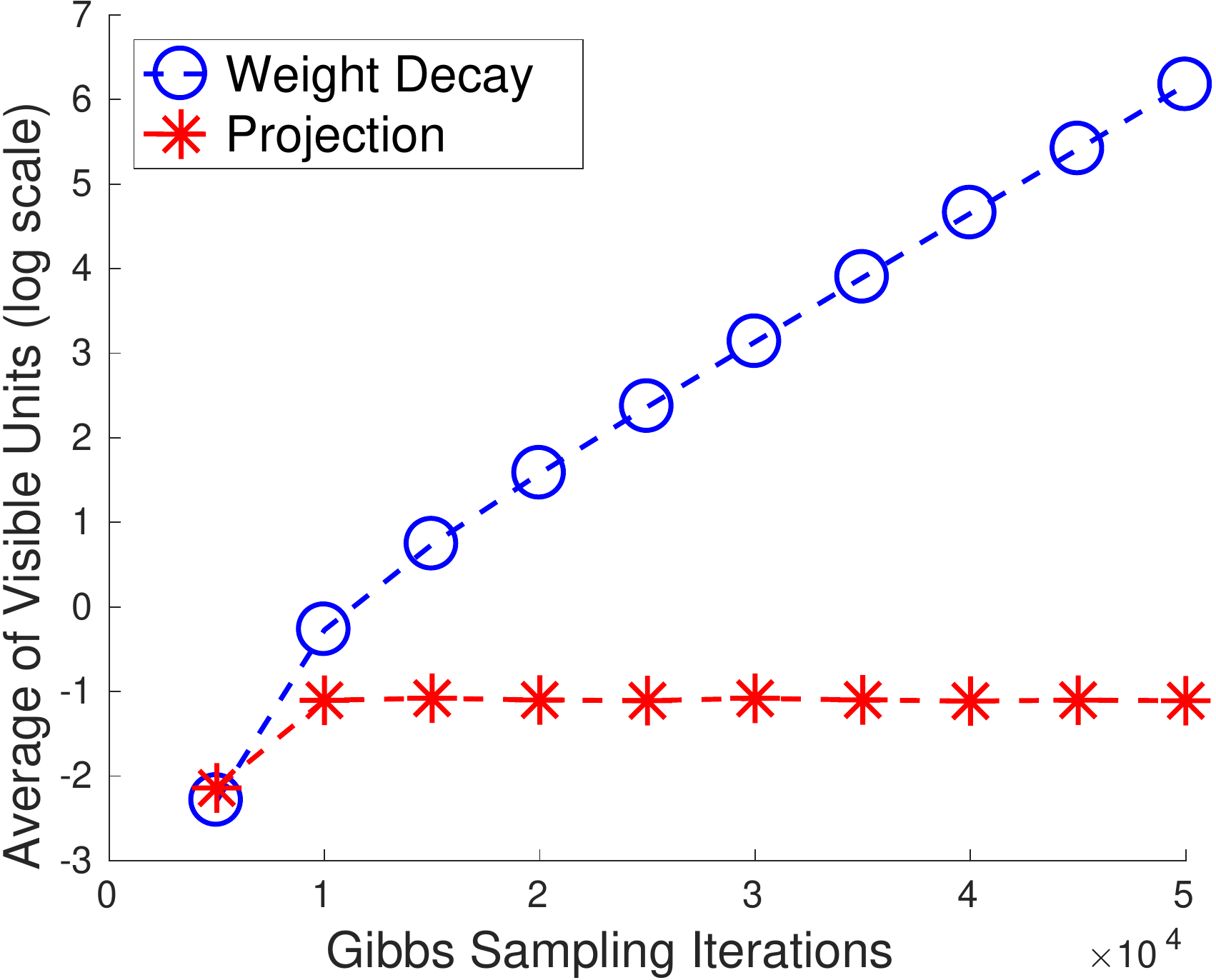}} 
  \caption{Divergence results on two datasets.}
  \label{fig:div}
\end{figure}

We conduct a short comparison to demonstrate the projection step is necessary for the leaky RBM on generative
tasks.  
We train two leaky RBM as follows. The first model is trained by the same setting in Section~\ref{sec:par}. 
We use the convergence of log likelihood as the stopping criteria. 
The second
model is trained by CD-1 with weight decay and without the projection step. We stop the training when the reconstruction
error is less then $10^{-2}$. 
After we train these two models, we run Gibbs sampling with 1000 independent chains for several steps and output the 
average value of the visible units. Note that the visible units are normalized to zero mean.
The results on SVHN and CIFAR10 are shown in Figure~\ref{fig:div}.

From Figure~\ref{fig:div}, the model trained by weight decay without projection step is suffered by the problem of the 
diverged values. It confirms the study shown in Section~\ref{sec:truncate}.
It also implies that we cannot train leaky RBM with larger CD steps when we do not do projection; otherwise, we would
have the diverged gradients.
Therefore, the projection is necessary for training leaky RBM for the generative purpose. 
However, we also oberseve that the projection step is not necessary for the classification and reconstruction tasks.
he reason may be the
independency of different evaluation criteria~\citep{Hinton12, Theis2016a} or other implicit reasons to be studied.

\section{Equivalence between Annealing the Energy and Leakiness}
\label{sec:equiv}
We analyze the performance gap between AIS-Leaky and AIS-Energy. One major difference is the initial distribution.
The intermediate marginal distribution of AIS-Energy has the following form:
\begin{equation}
	p_k(v) \propto \displaystyle\exp\left( -\frac{1}{2}v^\top\left( I-(1-\beta_k)\sum_{\eta_j>0}W_jW_j^\top -
	(1-\beta_k)c\sum_{\eta_j\leq 0}W_jW_j^\top \right)v  \right).
	\label{eq:energy}
\end{equation}
Here we eliminated the bias terms $b$ for simplicity.  Compared with Algorithm~\ref{algo:sample}, \eqref{eq:energy}  not
only anneals the leakiness $(1-\beta_k)c\sum_{\eta_j\leq 0}W_jW_j^\top$ when $\eta_j\leq 0$, but also in the case
$(1-\beta_k)\sum_{\eta_j>0}W_jW_j^\top$ when $\eta_j> 0$, which brings more bias to the estimation.
In other words, AIS-Leaky is a \emph{one-sided leakiness annealing} 
while AIS-Energy
is a \emph{two-sided leakiness annealing} method.

To address the higher bias problem of AIS-Energy, we replace the initial distribution with the one used in
Algorithm~\ref{algo:sample}. By elementary calculation, the marginal distribution becomes 
\begin{equation}
	p_k(v) \propto \displaystyle\exp\left( -\frac{1}{2}v^\top\left( I-\sum_{\eta_j>0}W_jW_j^\top -
	(\beta_k+(1-\beta_k)c)\sum_{\eta_j\leq 0}W_jW_j^\top \right)v  \right),
	\label{eq:energy2}
\end{equation}
which recovers the proposed Algorithm~\ref{algo:sample}.
From this analysis, we understand AIS-Leaky is a special case of conventional
AIS-Energy with better initialization inspired by the study in Section~\ref{sec:relu}.
Also, by this connection between AIS-Energy and AIS-Leaky, we note that AIS-Leaky can be combined with other
extensions of AIS~\citep{GrosseMS13, BurdaGS15} as well.

\end{document}